\documentclass[10pt,leqno]{amsart}

\usepackage{iftex}

\usepackage[utf8]{inputenc}
\usepackage[T1]{fontenc}
\usepackage[british]{babel}

\usepackage{graphicx}
\usepackage{xcolor,paralist,hyperref,fancyhdr,etoolbox}
\usepackage{indentfirst,csquotes}
\usepackage{amssymb,amsthm,amsmath}
\usepackage{multicol}
\usepackage{booktabs}

\hypersetup{
  colorlinks=true,
  linkcolor=black,
  filecolor=black,
  urlcolor=black
}

\begin{document}
\title[Automating Thematic Review of PFD Reports]{Automating Thematic Review of Prevention of Future Deaths Reports: Replicating the ONS Child Suicide Study using Large Language Models}
\author[Osian, Dutta, Bhandari, Buchan, Joyce]{
  Sam Osian\textsuperscript{1,2,3} \and
  Arpan Dutta\textsuperscript{1} \and
  Sahil Bhandari\textsuperscript{1} \and
  Iain E. Buchan\textsuperscript{1,2} \and
  Dan W. Joyce\textsuperscript{1,2,3}
}

\date{\today}
\email{samoand@liverpool.ac.uk}

\let\thefootnote\relax

\maketitle

\begin{center}
\footnotesize
\textsuperscript{1} Mental health Research for Innovation Centre (M-RIC), Mersey Care NHS Foundation Trust and the University of Liverpool, Liverpool Science Park, 131 Mount Pleasant, Liverpool L3 5TF\\
\textsuperscript{2} Civic Health Innovation Labs, University of Liverpool, Liverpool Science Park, 131 Mount Pleasant, Liverpool L3 5TF\\
\textsuperscript{3} Department of Primary Care and Mental Health, Institute of Population Health, Waterhouse Building Block B, University of Liverpool, L69 3GF\\
\end{center}


\begin{abstract}
Prevention of Future Deaths (PFD) reports, issued by coroners in England and Wales, flag systemic hazards that may lead to further loss of life. Analysis of these reports has previously been constrained by the manual effort required to identify and code relevant cases. In 2025, the Office for National Statistics (ONS) published a national thematic review of child-suicide PFD reports ($\leq$ 18 years), identifying 37 cases from January 2015 to November 2023 — a process based entirely on manual curation and coding. We evaluated whether a fully automated, open source ``text-to-table'' language-model pipeline (PFD Toolkit) could reproduce the ONS’s identification and thematic analysis of child-suicide PFD reports, and assessed gains in efficiency and reliability. All 4,249 PFD reports published from July 2013 to November 2023 were processed via PFD Toolkit's large language model (LLM) extraction and coding pipelines. Automated screening identified cases where the coroner attributed death to suicide in individuals aged 18 or younger, and eligible reports were coded for recipient category and 23 concern sub-themes, replicating the ONS coding frame. PFD Toolkit identified 72 child-suicide PFD reports — almost twice the ONS count. Three blinded clinicians adjudicated a stratified sample of 144 reports to validate the child-suicide screening. Against the post-consensus clinical annotations, the LLM-based workflow showed substantial to almost-perfect agreement (Cohen's $\kappa$ = 0.82, 95\% CI: 0.66-0.98, raw agreement = 91\%). The end-to-end script runtime was 8 minutes and 16 seconds, transforming a process that previously took months into one that can be completed in minutes. This demonstrates that automated LLM analysis can reliably and efficiently replicate manual thematic reviews of coronial data, enabling scalable, reproducible, and timely insights for public health and safety. The PFD Toolkit is openly available for future research.

\end{abstract} 

\bigskip

\begin{multicols}{2}

\section{Background}

Prevention of Future Deaths (PFD) reports are statutory notices from coroners in England and Wales, highlighting circumstances that may risk further deaths. Each PFD report describes the context of an inquest, details factors believed to have contributed, and lists concerns that the coroner believes the recipients of the reports are well-placed to address.

These documents represent a unique window into ground-level public safety hazards. However, despite being publicly available on the \textit{judiciary.uk} website \cite{judiciaryPFD}, the research potential of PFD reports has remained largely underexploited, primarily due to practical barriers to access and analysis. Reports are dispersed across a poorly indexed judicial website, lack consistent metadata, and are often available only as scanned images which lack searchable text. No downloadable dataset is available from \textit{judiciary.uk} for systematic retrieval of report content. Although the online repository provides category-based filters (such as care home deaths, deaths in hospitals, etc.), these are inconsistently applied; around 70\% of reports lack any form of category-based tag altogether \cite{zhang_lessons_2023}.

Consequently, even high-profile research must rely on laborious manual screening and annotation, demanding months or even years of researcher capacity \cite{bremner_systematic_2023, wallace_thematic_2024}. The House of Commons Justice Committee conceded in 2021 that this system is ``under-developed'' for public safety, citing the lack of ability to search and analyse reports to identify recurring issues \cite{justice_committee}.

The Office for National Statistics (ONS) has offered the first systematic study of PFD reports related to child suicide. Its 2025 bulletin involved the manual screening of hundreds of PFD reports, identifying 37 cases related to child suicide published between January 2015 and November 2023. The ONS researchers hand-coded reports by recipient, as well as 23 `coroner concerns' topics across 6 broad themes, from service-provision failures to gaps in communication and staffing \cite{ons2025}. 

To address the manual burden placed upon researchers in analysing PFD reports (as exemplified by the ONS study), there is a need for tools that can reliably and efficiently process the corpus, including scanned or untagged reports, without relying on inconsistent manual filters or metadata. An ideal solution would provide end-to-end automation of tasks that currently require manual effort from researchers, including screening for relevant cases, discovering themes, and generating user-defined metadata (e.g. age of the deceased).

Recent advances in large language models (LLMs), including multi-modal models capable of image ingestion, offer a way to overcome these longstanding constraints. We developed the PFD Toolkit, an open source Python package (\url{https://github.com/Sam-Osian/PFD-toolkit}) \cite{pfdtoolkit}, for automated extraction of structured data from these unstructured PFD reports (i.e. ``text-to-table''). The PFD Toolkit integrates LLMs and optical character recognition (OCR) web scraping, along with LLM-driven coding, to enable systematic interrogation of the entire PFD corpus, regardless of file format or quality. This automated approach enables highly scalable and reproducible research pipelines, in contrast to the lengthy and opaque processes previously required.

By applying PFD Toolkit’s automated Vision-LLM pipeline to the entire corpus of 4,249 PFD reports (July 2013 – November 2023), we replicate the ONS coding frame at scale, recover missed cases, and benchmark performance against post-consensus clinical annotations. Through this, we offer a comprehensive and reproducible alternative to months-long manual review.

This work complements existing initiatives such as the Preventable Deaths Tracker (PDT) \textit{(\url{https://preventabledeathstracker.net/})}, which enhances access and searchability of PFD reports. Unlike the PDT, which deals only in PFD report metadata (such as location, date, and recipients), PFD Toolkit processes both metadata and the long-text content of each report. It also provides an end-to-end automated workflow for user-defined screening and thematic coding of these reports, supporting scalable and reproducible research applications.

\section{Objectives}

This study sought to assess whether a fully automated pipeline can replicate the ONS manual review of child-suicide PFD reports, with respect to case identification, thematic coding, and efficiency. We addressed three research questions:

1. How does the number of child-suicide PFD reports identified by an LLM-based pipeline compare with those identified by the ONS manual review?

2. To what extent does the pipeline’s thematic analysis agree with independent clinical adjudication?

3. How does the time required for automated identification and coding compare with the time required for manual review?
$\,$

\end{multicols}

\begin{figure}[htbp]
  \centering
  \includegraphics[width=0.8\textwidth]{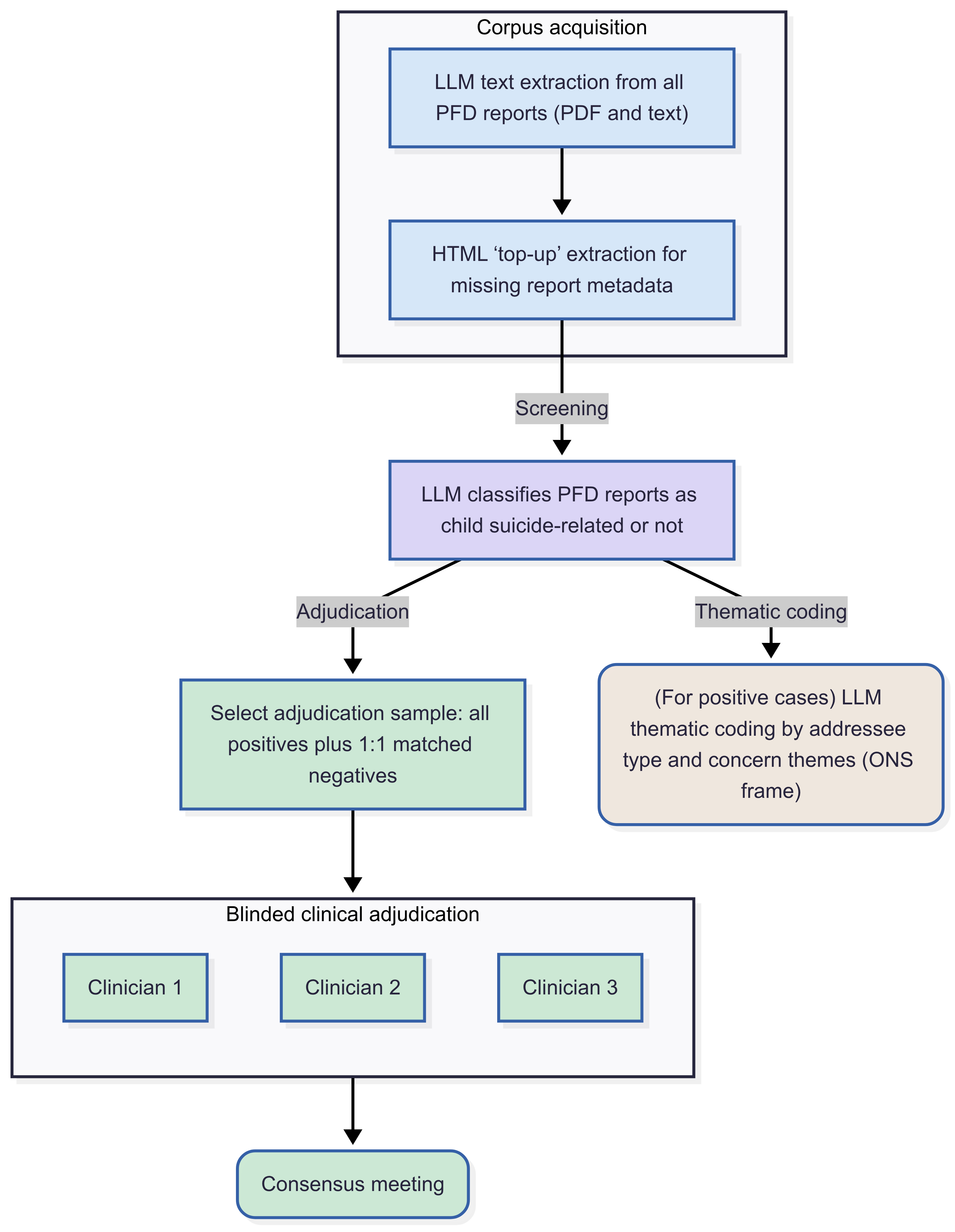} 
  \caption{Workflow diagram for corpus acquisition, screening, thematic coding, and clinical adjudication stages of the study.}
  \label{fig:workflow}
\end{figure}
\begin{multicols}{2}

\section{Methods}

\subsection{Study design}

We conducted a concordance study, comparing a fully automated language-model pipeline to the ONS’s manual review of child-suicide PFD reports. The pipeline, implemented in PFD Toolkit v0.3.3, handled all analytical stages: ingestion, screening, coding, and tabulation. Outputs were benchmarked against two reference standards:

1. The total number of reports identified in the published ONS study ($n = 37$).

2. An independent, blinded clinical review of a stratified, decoy-boosted sample of 144 reports, used as a gold-standard for accuracy/agreement assessment.

Primary outcomes were: (i) number of child suicide PFD reports identified, (ii) kappa agreement for identified reports against a clinical gold standard, and (iii) end-to-end script runtime for the entire analysis. 

Unfortunately, the ONS did not release its full post-screen report list, meaning we can only compare report identification with the ONS study at an aggregate level.

\subsection{Corpus acquisition}

The PFD Toolkit scraped all reports under the `prevention-of-future-death-reports/' path from \textit{https://judiciary.uk/}. The toolkit processed both text-based and scanned documents using a multi-stage pipeline that includes 200 dpi Base64 image conversion and vision-enabled LLM text extraction via OCR, followed by HTML scraping as a fail-safe to retrieve any missing metadata.

All 4,249 unique reports published from July 2013 to November 2023 were included in the initial corpus acquisition for this analysis. Notably, the ``suicide'' and ``child death'' tags were only introduced on to the judiciary.uk website in January 2015 and were not applied to reports retrospectively, limiting the ONS study to reports published from January 2015 onwards. By using full-text and image-based screening (as an alternative to judiciary.uk's category-based searches), PFD Toolkit is able to identify relevant cases across the entire available time frame (i.e. from the system's inception in July 2013). Our upper time window (November 2023) matches the ONS frame.

\subsection{Screening and thematic coding}

We prompted \textit{GPT 4.1} \cite{gpt} to code reports by:

1. Relevance to child suicide. As per the ONS frame, this was inclusive of those aged 18. We also allowed the model to infer child status from written \textit{cues} (e.g. mentioning recent ``Child and Adolescent Mental Health Services'' (CAMHS) usage, being in ``Year 10'', etc.).

2. Addressee categories: NHS body; Government department/minister; Local authority; Professional body; Other. The ONS did not define `professional body', and so our research adopted the following definition: \textit{``an organisation with statutory responsibility for a profession (GMC, NMC, Royal Colleges, etc.)''}.

3. Coroner concerns: 23 sub-themes under 6 headline themes as per the ONS coding frame: Service provision; Staffing \& resources; Communication; Multi-service care, Accessing services, Access to harmful content \& environment.

For each stage, outputs were coerced into JSON arrays containing boolean values, where \textit{True} represented a positive case for a given report, and \textit{False} represented a negative case. Pydantic \cite{pydantic} was used to enforce and validate this output schema.

To support transparency, the toolkit can optionally output supporting text evidence for each coded item, helping users to audit decisions and facilitate interpretability. Although automated, the toolkit's approach therefore remains open to human-in-the-loop workflows or post hoc review as needed.

\subsection{Clinical adjudication}

To benchmark the accuracy of the automated approach as per \#1 above, the lead researcher \textit{(S.O.)} assembled a stratified, decoy-boosted sample of 144 reports: 72 flagged as positive by the toolkit (i.e. related to child suicide) and 72 as negative. Including a balanced mix of positive and negative cases was intended to prevent reviewers from inferring the expected distribution of cases, thereby reducing the risk of response bias. Three clinicians \textit{(D.J., A.D., and S.B.)} provided a ground-truth data set, rating each of the 144 reports as either a child suicide, or not.  All three clinicians have membership of the Royal College of Psychiatrists, experience of serious incident reviews and the UK's coronial inquest system.  They were blind to the algorithm's outputs, to one another’s assessments and independently classified each report. Where the clinicians disagreed, discrepancies were resolved through a consensus meeting, resulting in a final set of gold-standard labels for comparison.  For ambiguous cases, the clinicians were instructed to use the `balance of probability' standard for judging if the report related to a suicide; for example, if suicidal intent was not established and explicitly recorded in the PFD report, clinicians were asked to review the PFD report text and decide.

The number of clinical reviewers was determined using the fixed-n binary kappa power analysis function from R's `kappaSize' package. With an anticipated kappa of 0.70, the estimated lower bound of the 95\% confidence interval was estimated as 0.589 for two reviewers, 0.617 for three, 0.626 for four, and 0.630 for five. Since three reviewers exceeded the conventional ``substantial agreement'' threshold ($\geq$0.61) \cite{kappa}, and additional reviewers offered only minor improvements, a panel of three was deemed sufficient for this research.

\subsection{Computational environment}

All analyses were run on a consumer-grade Ubuntu 24.04 laptop, with LLM calls made to \textit{GPT 4.1} handled by OpenAI’s cloud API. PFD Toolkit also allows for local LLM frameworks (e.g. Ollama) as an alternative to proprietary, server-side tooling.

\subsection{Ethics}

PFD reports are already in the public domain. Any sensitive content is redacted by the Chief Coroner's Office \cite{chief_coroner_pfd_policy}, meaning that no personally identifiable information of living persons was processed. On this basis, and with institutional ethical approval from the University of Liverpool (Institute of Population Health Ethics Review Board, Ref: 16296, 04/06/2025), research use of this data, as well as server-side processing via third-party APIs, was deemed appropriate.

\section{Results}

\subsection{Report identification}

(See Table~\ref{tab:identification}). The PFD Toolkit identified a total of 72 child-suicide PFD reports published between July 2013 and November 2023, compared with 37 reports identified in the ONS study between January 2015 and November 2023. Of the 72 identified by the Toolkit, 62 were published within the ONS's considered time frame, while 10 were published prior to the introduction of `suicide' and `child death' categories on the judiciary.uk website in January 2015.

\subsection{Clinical adjudication}

In the validation sample ($n = 144$), inter-rater reliability among the three clinical reviewers prior to consensus adjudication was [Fleiss' $\kappa$ = 0.766, 95\% CI: 0.672–0.861], suggesting substantial to almost-perfect agreement.\textit{ (According to the widely cited Landis and Koch scale, kappa values between 0.61 and 0.80 can be interpreted as ``substantial agreement,'' while values above 0.80 indicate ``almost perfect agreement.'' \cite{kappa}).}

Post-consensus adjudication, agreement between the PFD Toolkit output and the clinical reference standard was substantial to almost-perfect (Cohen's $\kappa$ = 0.82, 95\% CI: 0.66-0.98, raw agreement = 91\%). 

\end{multicols}

\begin{table}[htbp]\centering
\caption{Comparing PFD Toolkit and the ONS in identifying child‐suicide PFD reports}
\begin{tabular}{@{}lcc@{}}
\toprule
 & \textbf{PFD Toolkit}& \textbf{ONS} \\ \midrule
ONS window (Jan 2015 – Nov 2023)& 62 & 37 \\
Pre-2015& 10& -- \\
 \textbf{Total}& \textbf{72}&\textbf{37}\\
\end{tabular}
\label{tab:identification}
\end{table}

\begin{multicols}{2}

\end{multicols}

\begin{table}[htbp]\centering
\caption{Addressee categories coded by PFD Toolkit, based on the ONS coding frame ($n = 72$ reports).}
\begin{tabular}{@{}lc}
\toprule
\textbf{Category} & \textbf{Count of reports}\\ \midrule
NHS Trust or CCG& 40  \\
Government department or minister& 33  \\
Local authority                    & 17  \\
 Professional body&8\\
Other&  25\\ \bottomrule
\end{tabular}
\label{tab:addressees}
\end{table}

\begin{table}[htbp]\centering
\caption{Coroner‐concern sub-themes coded by PFD Toolkit, based on the ONS coding frame ($n = 72$ reports).}
\begin{tabular}{@{}lc}
\toprule
\textbf{Theme / sub-theme}& \textbf{Count of reports}\\ \midrule
	\multicolumn{2}{l}{\textbf{Service provision}}\\
 Standard operating procedures
or processes not followed
or adequate&41\\
Risk assessment	& 27 \\
	Specialist services
(crisis, autism, beds)& 20 \\
Discharge from services	&  7 \\
Diagnostics&  9 \\[2pt]
 \multicolumn{2}{l}{\textbf{Staffing and resources}}\\
	Training missing, inadequate
or not mandatory& 28 \\
Inadequate staffing 	& 17 \\
Lack of funding 	& 14 \\
Recruitment and retention problems 	&  4 \\[2pt]
 \multicolumn{2}{l}{\textbf{Communication}}\\
Lack of communication
between services& 31 \\
Within services
communication is poor& 17 \\
Lack of communication
with patient and family& 15 \\
	Confidentiality risk
not communicated&  8 \\[2pt]
 \multicolumn{2}{l}{\textbf{Multiple services involved in care}}\\
	Issues with Local Authority
(including child services, schools)& 16 \\
	Integration of care
was disconnected& 14 \\
Transition from CAMHS to
adult services ineffective&  7 \\[2pt]
 \multicolumn{2}{l}{\textbf{Accessing services}}\\
Delays in referrals
and waiting times& 21 \\
	Patient
engagement
lacking& 10 \\
	Referrals rejected &  7 \\[2pt]
 \multicolumn{2}{l}{\textbf{Access to harmful content and environment}}\\
	Access to harmful
items or substances&  6 \\
	Internet content
and controls&  4 \\
Access to Trainlines 	&  3 \\
	Safeguarding from
sensitive material&  2 \\ \bottomrule
\end{tabular}
\label{tab:themes}
\end{table}

\begin{multicols}{2}

\subsection{Run-time}

The end-to-end processing (wall) time for the workflow was 8 minutes and 16 seconds. This included querying the entire corpus ($n = 4,249$) for cases relevant to child suicide, in addition to thematic coding and results tabulation.

\subsection{Thematic outputs}

Table~\ref{tab:addressees} and Table~\ref{tab:themes} mirror the tables produced as part of the ONS analysis. Because a single report may have multiple addressees or concern sub-themes, the totals in each table are greater than the total number of reports analysed.

Importantly, the ONS presented counts of coroner concerns based on the number of ``mentions'' within and across reports, meaning that a single report could contribute multiple counts to each sub-theme or addressee category. In contrast, PFD Toolkit results are presented at the report level (i.e.\ each report is counted once per sub-theme, regardless of how many times a concern is mentioned within it). As such, the figures in Table~\ref{tab:themes} cannot be directly compared to the corresponding figures in the ONS bulletin.

\section{Conclusions}

This study demonstrates that a fully automated, LLM-based workflow can reliably deliver efficient thematic reviews of Prevention of Future Deaths reports at scale — transforming a process that previously took months into one completed in just over 8 minutes. The PFD Toolkit identified more child-suicide PFD reports than the ONS manual review (72 vs. 37) and applied the identical coding frame to all identified cases. The pipeline achieved substantial to almost-perfect agreement with independent clinical review in identifying relevant cases. This approach enables transparent, reproducible, and rapid analysis of coronial data, supporting future research and surveillance.

Beyond the present case study, PFD Toolkit is designed as a generalisable infrastructure for all forms of research and policy work involving PFD report data. Its modular, domain-agnostic architecture allows users to define and extract any set of variables, concerns, or risk factors relevant to their research question — from deaths involving restraint or medication safety, to patterns in care home incidents or maternal mortality. By minimising the need for manual review, the toolkit lowers the barriers to research on preventable deaths, enabling researchers, audit teams, and public agencies to rapidly conduct systematic reviews or surveillance on any topic of emerging concern. This flexibility positions PFD Toolkit as a foundation for a wide range of future analyses, enabling a step-change in the scale, speed, and reproducibility of research using Prevention of Future Deaths reports.

In addition, PFD Toolkit supports automated discovery of emergent themes, as well as concise, user-friendly report summaries to facilitate audit and policy work. Users can also select from a range of language models, ensuring the toolkit’s accuracy and versatility will continue to improve as LLM technology evolves.  

The software and example analysis notebooks are openly available to support transparency and further methodological development (\url{https://github.com/Sam-Osian/PFD-toolkit}).

\subsection{Limitations}

Coroners are not required to record the age of the deceased in PFD reports, and in practice, age may only be included when considered contextually relevant. While we believe that child status is likely to be documented in most suicide cases, it remains possible that some relevant cases were missed if the coroner did not specify the deceased’s age. However, this is a data quality/completeness issue, rather than a limitation of PFD Toolkit itself. 

The performance of PFD Toolkit is also likely to depend on the specific language model used for extraction and coding. This research used OpenAI’s \textit{GPT 4.1} via the cloud API. Different LLMs may yield varying levels of accuracy and efficiency, and future updates or alternative models could produce different results.

Because the ONS did not release their annotated report-level dataset, we were unable to directly compare our screening results on a report-by-report basis. Additionally, the ONS counted each mention of a concern within and across reports, whereas our approach reports the presence of each concern at the report level only. This difference in counting methods limits the comparability of sub-theme frequencies between the two analyses.

\section*{Conflict of interest}

The authors declare no conflicts of interest.

\section*{Funding}
S.O. is funded by a philanthropic gift to the University of Liverpool from Sir Robin Saxby.  I.E.B. and D.W.J. are in part supported by the National Institute for Health and Care Research (NIHR) Mental Health Translational Research Collaboration, hosted by the NIHR Oxford Health Biomedical Research Centre.

\bibliographystyle{unsrt}  
\bibliography{main}

\end{multicols}

\end{document}